%% file: cvpr.tex
\documentclass[final]{cvpr}
\pdfoutput=1 
\usepackage{times}
\usepackage{epsfig}
\usepackage{graphicx}
\usepackage{amsmath}
\usepackage{amssymb}
\usepackage{pifont}
\usepackage{bm}
\usepackage{subcaption}
\usepackage{booktabs}
\usepackage{colortbl}
\usepackage[dvipsnames]{xcolor}
\usepackage [english]{babel}
\usepackage [autostyle, english = american]{csquotes}
\usepackage{comment}

\MakeOuterQuote{"}

\newcommand{\COUNTIMAGES}{106K}
\newcommand{\COUNTCATEGORIES}{15}
\newcommand{\INDOFASHION}{IndoFashion}

\definecolor{ashgrey}{rgb}{0.7, 0.75, 0.71}
\makeatletter
\newcommand\footnoteref[1]{\protected@xdef\@thefnmark{\ref{#1}}\@footnotemark}
\makeatother

\usepackage[breaklinks=true,bookmarks=false]{hyperref}

\def\cvprPaperID{14} 

\begin{document}

\title{IndoFashion : Apparel Classification for Indian Ethnic Clothes}

\author{Pranjal Singh Rajput\\
Delft University of Technology\\
{\tt\small pranjal.tudelft@gmail.com}
\and
Shivangi Aneja\\
Technical University of Munich\\
{\tt\small shivangi.aneja@tum.de}
}


\twocolumn[{%
\renewcommand\twocolumn[1][]{#1}%
\maketitle
\begin{center}
    \centering
    \includegraphics[width=1.0\linewidth]{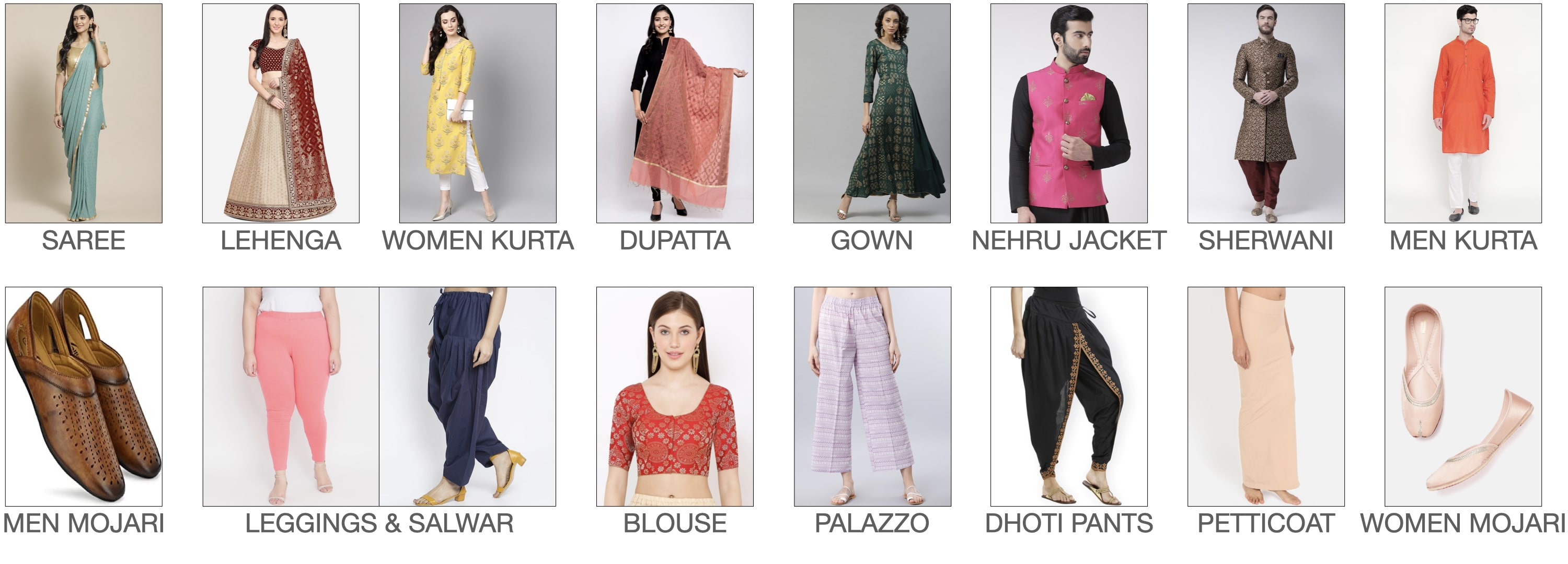}
    \captionof{figure}{This paper proposes a new dataset with \COUNTIMAGES{} images and \COUNTCATEGORIES{} classes for cloth categorization for Indian style clothing. Figure shows images from the dataset with the cooresponding labels. }
    \label{fig:teaser}
\end{center}%
}]

\input{pages/00_abstract}
\input{pages/01_intro}

\input{pages/02_related_work}
\input{pages/03_dataset}
\input{pages/04_experiments}
\input{pages/05_results}

\input{pages/06_conclusion}

{\small
\bibliographystyle{ieee_fullname}
\bibliography{egbib}
}

\end{document}

%% file: pages/00_abstract.tex
\begin{abstract}
Cloth categorization is an important research problem that is used by e-commerce websites for displaying correct products to the end-users.  Indian clothes have a large number of clothing categories both for men and women. The traditional Indian clothes like "Saree" and "Dhoti" are worn very differently from western clothes like t-shirts and jeans. Moreover, the style and patterns of ethnic clothes have a very different distribution from western outfits. Thus the models trained on standard cloth datasets fail miserably on ethnic outfits. To address these challenges, we introduce the first large-scale ethnic dataset of over \COUNTIMAGES{} images with \COUNTCATEGORIES{} different categories for fine-grained classification of Indian ethnic clothes. We gathered a diverse dataset from a large number of Indian e-commerce websites. We then evaluate several baselines for the cloth classification task on our dataset. In the end, we obtain 88.43 \% classification accuracy. We hope that our dataset would foster research in the development of several algorithms such as cloth classification, landmark detection, especially for ethnic clothes.
\end{abstract}

%% file: pages/01_intro.tex
\section{Introduction}

With the increasing use of online shopping for clothing, it has become important for e-commerce websites to correctly identify categories for all the clothing items available on the website. Few datasets like FashionMNIST~\cite{Xiao2017FashionMNISTAN}, DeepFashion~\cite{Ge2019DeepFashion2AV, Liu2016DeepFashionPR} have been introduced in past few years to identify cloth categories. However, these datasets mostly deal with western attire like shirts, jeans, jackets, shorts, etc. In contrast, our focus in the work is on traditional Indian attire where the cloth categories like Saree, Kurta, Dhoti, etc are very different from western ones, see Figure~\ref{fig:teaser}. With this work, we introduce an ethnic fashion dataset, a first of its kind to identify these cloth categories. 
\\
\\
\noindent
To the best of our knowledge, only two datasets Atlas~\cite{atlas} and Generative Indian Fashion~\cite{genfashion} deal with Indian ethnic clothes. Although Atlas~\cite{atlas} dataset deals with identifying cloth taxonomy, the dataset is highly imbalanced with less than 40 images for few class categories, make it unsuitable for training deep models. Generative Indian Fashion~\cite{genfashion} does not address the cloth classification problem and has very few images (12K) and class categories. Thus, both these datasets are limited in terms of either dataset size or applicability for cloth classification of Indian ethnic wear. 
\\
\\
\noindent
We notice that no benchmarking is done for cloth categorization of Indian Ethnic clothes. In this work, we propose to address the problem by providing the first large-scale dataset for the task. In addition, we also benchmark several state-of-the-art image classification models to evaluate the performance on the dataset.
\\
\\
\noindent
In summary, our contributions are as follows:
\begin{itemize}
  \item To the best of our knowledge, this paper proposes the first automated method to categorize Indian Ethnic clothes.
  \item We created a large dataset of \COUNTIMAGES{} images with \COUNTCATEGORIES{} different cloth categories from a variety of e-commerce websites.
  \item We benchmark our results on several image classification results and analyze their performance.
\end{itemize}


%% file: pages/02_related_work.tex
\section{Related Work}

\subsection{Clothing Datasets}
Several datasets ~\cite{Ge2019DeepFashion2AV, Liu2016DeepFashionPR, Xiao2017FashionMNISTAN, fashionAI} have been proposed in the recent past for clothes recognition. FashionMNIST~\cite{Xiao2017FashionMNISTAN} provides a toy dataset of 70K greyscale images of 10 cloth categories, lacking real-world scenarios. DeepFashion~\cite{Ge2019DeepFashion2AV} provides a very large-scale dataset with dense annotations for a wide variety of cloth-related research problems like cloth category/texture/fabric classification, clothing landmarks, etc. However, the dataset does not contain ethnic clothes. We, however, specifically target Indian ethnic clothes.

\subsection{Indian Cloth Datasets}

Only a few datasets~\cite{atlas, genfashion} are available, that contain the Indian ethnic clothes. Atlas~\cite{atlas} contains in total 186k clothing images focusing on product taxonomy classification, consisting of images corresponding to western wear, inner-wears, and ethnic wears. However, the dataset is very imbalanced in the case of some cloth categories. In Generative Indian Fashion~\cite{genfashion}, the authors focus on modifying the clothing of a person according to a textual description. The dataset is very small consisting of only 12k images. Also, the authors do not address the problem of cloth categorization. On the other hand, \INDOFASHION{} dataset consists of \COUNTIMAGES{} images and \COUNTCATEGORIES{} categories, providing a rich, diverse and balanced dataset, that focuses on ethnic fashion research.

%% file: pages/03_dataset.tex
\section{IndoFashion Dataset}

\begin{figure}[!ht]
\begin{center}
 \includegraphics[width=1.0\linewidth]{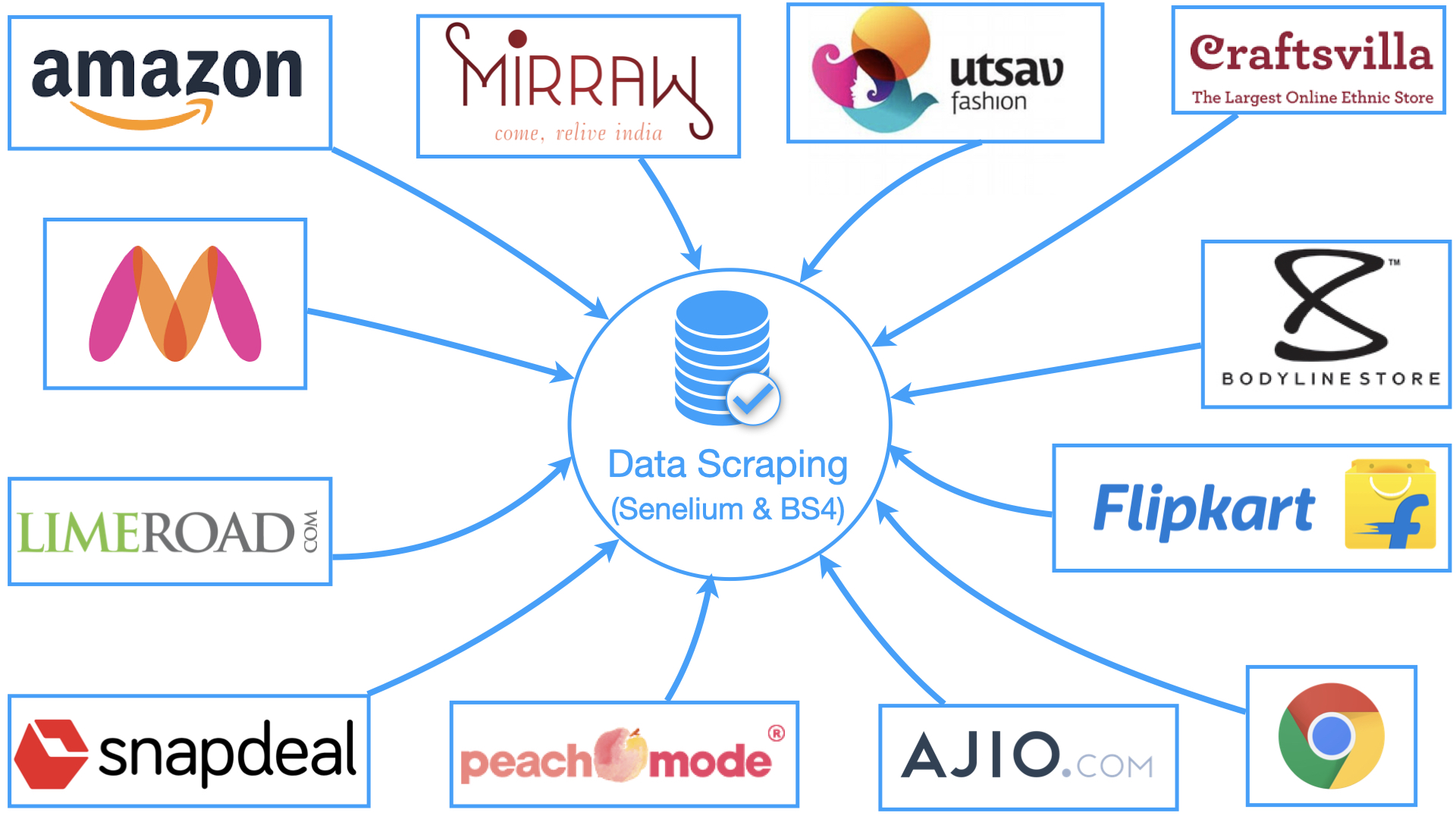}
\end{center}
   \caption{For our dataset, we gathered images from e-commerce websites and google image search as shown in figure. To scrape the images custom webscapers based on beautifulsoup~\cite{richardson2007beautiful} and selenium webdriver~\cite{selenium} were used.}
\label{fig:data_collection}
\end{figure}

\begin{figure}[!ht]
\begin{center}
 \includegraphics[width=1.0\linewidth]{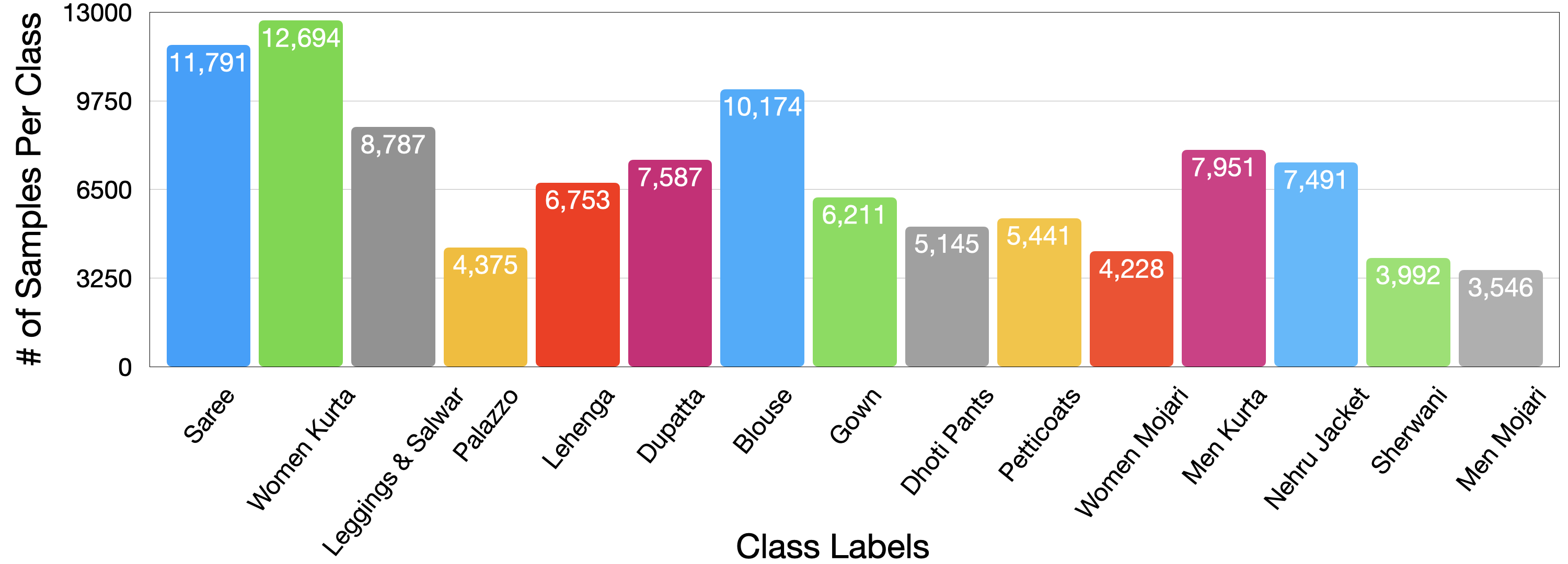}
\end{center}
   \caption{Figure shows category-wise frequency distribution of the 15 cloth categories in our proposed dataset.}
\label{fig:class_dist}
\end{figure}

IndoFashion is a unique fashion dataset consisting of a variety of Indian ethnic clothes. It is one of its kind of dataset and to the best of our knowledge, this is the first large-scale dataset for ethnic cloth classification. It contains \COUNTIMAGES{} images of \COUNTCATEGORIES{} different categories covering the most famous Indian ethnic clothes. The images are collected from google image search and several Indian e-commerce websites ~\cite{ajio, amazon, bodylinestore, craftsvilla, flipkart, limeroad, mirraw, myntra, peachmode, snapdeal, utsavfashion}, capturing different styles, colors, and patterns. Every image in the dataset is annotated with a label that represents its class category. The distribution of the dataset per category is shown in Figure~\ref{fig:class_dist}. Some example images per class label are shown in Figure ~\ref{fig:teaser}.

\subsection{Dataset Collection}
We gathered our dataset primarily from famous Indian e-commerce websites such Flipkart~\cite{flipkart}, Amazon~\cite{amazon}, Myntra~\cite{myntra}, etc, refer Figure \ref{fig:data_collection}. We curate our dataset in two steps: First, we used our custom scrapers to gather images corresponding to each of the categories in the dataset. Second, the dataset is cleaned by removing the duplicate images from the dataset. In the end, several images per class category is obtained with varying styles and colors.

\subsection{Dataset Statistics}
Our dataset consists of \COUNTIMAGES{} images and \COUNTCATEGORIES{} unique cloth categories. For the cloth classification task, we split the dataset into training, validation, and test set. For a fair evaluation, we ensure equal distribution of the classes in the validation and the test set consisting of 500 samples per class. We spent a total of 25 hours to collect, cleanup, and restructure the dataset. Our proposed dataset consists of 15 cloth categories as documented in Table~\ref{table:categories} and dataset split is shown in Table~\ref{tab:Dataset Split and stats}. 

\begin{table}[bht!]
\begin{center}
\begin{tabular}{  p{1cm}  p{6cm}}
\toprule
Gender & Categories\\ 
\midrule
Women & Saree, Women Kurta, Leggings \& Salwar, Palazzo, Lehenga, Dupatta, Blouse, Gown, Dhoti Pants, Petticoats, Women Mojari \\ 
\addlinespace
Men & Men Kurta, Nehru Jacket, Sherwani, Men Mojari\\ 
\bottomrule
\end{tabular}
\end{center}
\caption{The table lists the class categories for both men and women in the dataset.}
\label{table:categories}
\end{table}

\begin{table}[bht!]
\begin{center}
\begin{tabular}{c  c}
\toprule
Split & No. of Images  \\
\midrule
Train & 91,166 \\
Validation & 7,500 \\
Test & 7,500 \\
\bottomrule
\end{tabular}
\end{center}
\vspace{-0.4cm}
\caption{Statistics of our IndoFashion dataset.}
\label{tab:Dataset Split and stats}
\end{table}

\subsection{Comparison with other Indian ethnic datasets}

We compare our dataset against other existing Indian ethnic cloth datasets in Table~\ref{tab:datasetComparison}. Overall our dataset contains 106,166 images which is larger than any other existing dataset for ethnic clothes. Note that, Atlas dataset~\cite{atlas}, categorizes \textit{Men Suits} and \textit{Jackets} as ethnic wear, which are \textbf{not} considered ethnic clothes. Similarly, Jain \etal~\cite{genfashion} incorrectly categorize \textit{Casual Shirts}, \textit{Shirts}, \textit{T-Shirts} as ethnic wear. This incorrect categorization makes these datasets unsuitable for classification of ethnic clothes.

\begin{table}[bht!]
\begin{center}
\begin{tabular}{c c c c}
\toprule
Dataset & \#Images & \#Classes & Avg images\\
& & & per class\\
\midrule
Atlas dataset~\cite{atlas} & 81,813  & 17 & 4,812\\
Jain \etal~\cite{genfashion}  & 12,304 & 12 & 1,025 \\
\midrule
\textbf{\INDOFASHION{}(Ours)} & \textbf{106,166} & \textbf{\COUNTCATEGORIES{}} & \textbf{7,077}\\
\bottomrule
\end{tabular}
\end{center}
\vspace{-0.4cm}
\caption{Quantitative comparison of various Indian ethnic cloth datasets. \INDOFASHION{} dataset contains the largest number of images, in total as well as per category.}
\label{tab:datasetComparison}
\end{table}

%% file: pages/04_experiments.tex
\section{Experiments}

We evaluate several state-of-the-art Image Classification models for our task of classifying Indian ethnic clothes. All our models use an ILSVRC 2012-pretrained ResNet~\cite{resnet}-18, 50, and 101 backbone. All numbers reported in the paper are on the test split of the dataset. We use Precision, Recall, and F1-Score to measure the performance of our models.

\subsection{Pre-processing and Augmentations}\label{sec:aug}

We first pre-process all the images in the dataset by resizing the images to 128x256 pixels before feeding them into the network.

We also apply various types of augmentation to the images during training to mitigate over-fitting and improve the model performance. Note that we apply the augmentations to the training images only and are not applied during inference. We document these augmentations below:

\begin{enumerate}
    \item Geometric Transformation: Images are geometrically transformed by flipping them horizontally and rotating by a small angle of 10 degrees randomly. Images are geometrically transformed during training with a probability of 50\%. 
    \item Color Transformation: In this method, we apply color jitter by varying the hue and saturation by a factor of 0.2.
\end{enumerate}

\subsection{Hyperparameters}
We run all our experiments on an Nvidia GeForce GTX 2080 Ti card. We use a learning rate of 1e-3 with Adam optimizer for our model with decay when validation loss plateaus for 5 consecutive epochs, early stopping with the patience of 25 successive epochs on validation loss. 






%% file: pages/05_results.tex
\section{Results}
\begin{figure*}
\begin{center}
 \includegraphics[width=1.0\linewidth]{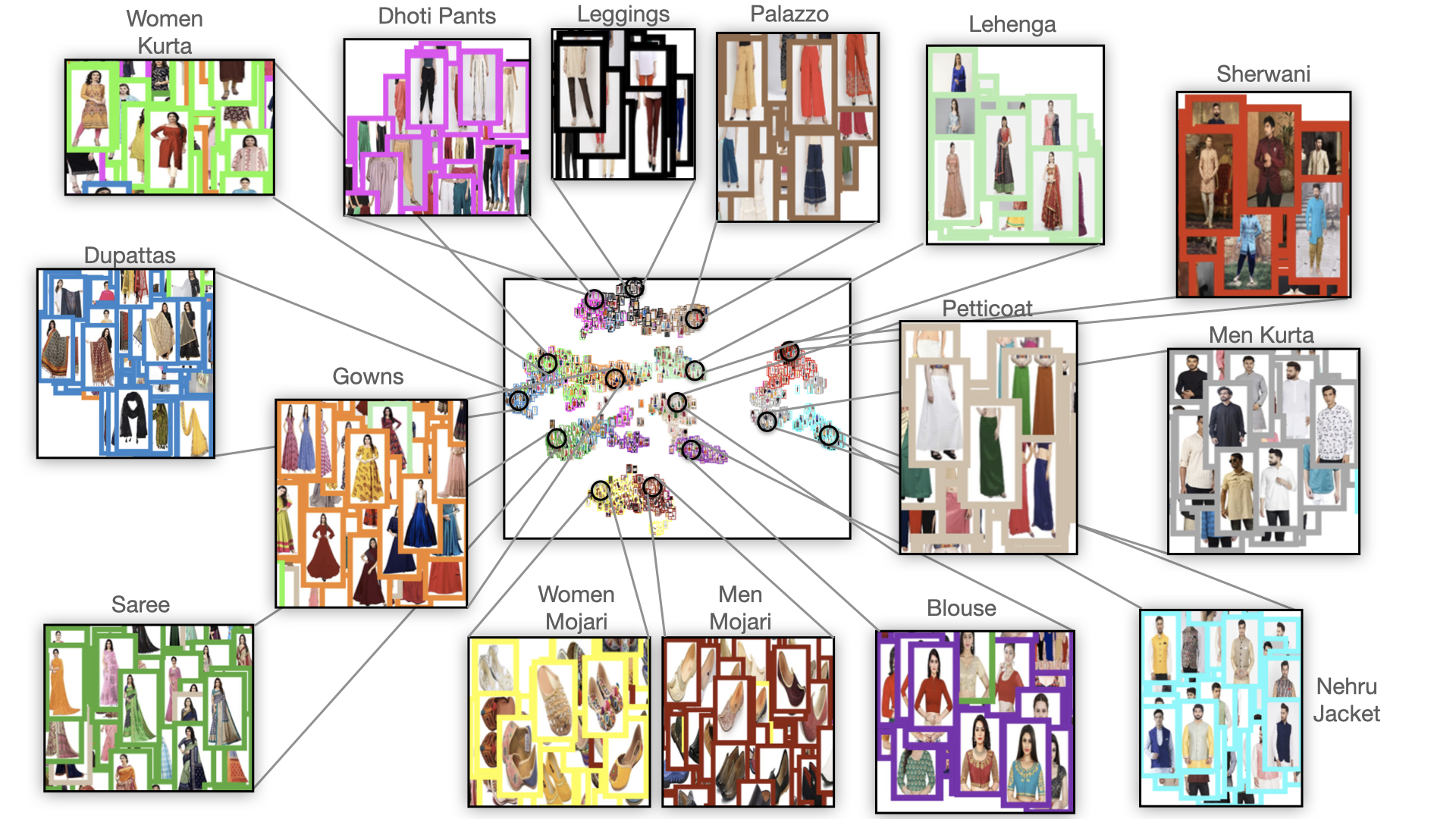}
\end{center}
  \caption{Figure shows the t-sne visualization of our best performing model. Each category is shown in a different color. Note that separate clusters are formed corresponding to each cloth category. }
\label{fig:t-sne}
\end{figure*}

\begin{figure}[!ht]
\begin{center}
 \includegraphics[width=1.0\linewidth]{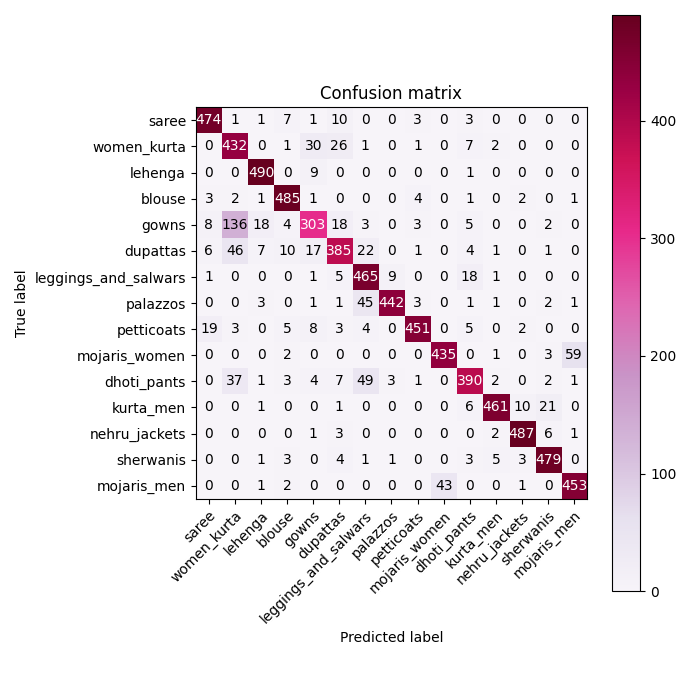}
\end{center}
\vspace{-0.5cm}
   \caption{Figure show the Confusion Matrix of our best performing model, Resnet-50(J+F), where (J+F)=Jitter and Flipping. Note that the class Women Kurta is often confused with Gowns, Mojari Men with Mojari Women since these pair of classes are visually similar to each other.}
\label{fig:best_model}
\end{figure}

We visualize the t-sne embedding in Figure \ref{fig:t-sne} and confusion matrix of our model in Figure \ref{fig:best_model}. In confusion matrix, it is observed that visually similar classes like \textit{Women Kurta} and \textit{Gowns} are difficult to distinguish using our trained model. This is also evident from the t-sne visualization, where the embeddings of these visually similar classes are close to each other.

\subsection{How Data Augmentation helps?}
We analyze the effect of augmentations mentioned in Section~\ref{sec:aug} in Table~\ref{tab:expResults1}. We plot the confusion matrix of our best performing model in Figure~\ref{fig:best_model}. We notice that applying all these augmentations together gives us the best performing model.

\begin{table}[bht!]
\begin{center}
\begin{tabular}{c c c c}
\toprule
Model & Precision & Recall & F1-Score\\
\midrule
R-18(NA)	& 86.9 & 86.12 & 86.11\\
R-18(F) & 88.3 & 87.41 & 87.44\\
R-18(J) & 87.83	& 87.24	& 87.24\\
R-18(J+F) & 88.62 & 87.52 & 87.46\\
\midrule
R-50(NA)	& 87.95	& 86.94	& 86.93\\
R-50(F) & 88.85 & 87.81 & 87.84\\
R-50(J) & 87.78 & 87.04 & 87.04\\
\textbf{R-50(J+F)} & \textbf{88.93}	& \textbf{88.42}	& \textbf{88.38}\\
\midrule
R-101(NA) & 87.66 & 86.82 & 86.86\\
R-101(F) & 89.35 & 88.56 & 88.5\\
R-101(J) & 86.86	 & 86.3 & 86.29\\
R-101(J+F) & 88.63 & 87.97 & 87.94\\
\bottomrule
\end{tabular}
\end{center}
\caption{The table illustrates the effect of applying different augmentations during training. To avoid clutter, following notations are used: R=ResNet, (J)=Jitter augmentation, (F)=Flipping, and (J+F)=Jitter and Flipping together, (NA)=No Augmentation.}
\label{tab:expResults1}
\end{table}

It is observed that the performance improves by implementing either geometrical or color augmentation alone. However, applying both augmentations together boosts the performance of the model. Also, we noticed that increasing the model size from resnet-18 to resnet-50 improves the performance further. However, increasing the model size to more bigger resnet-101 backbone does not show any improvements. Our best performing model is resnet-50 achieving a precision of 88.93\%. So we conclude that very large and complex models are not required for our dataset.

\subsection{How training data size affects?} 
We experiment with different percentage of training data size (w.r.t. our full data of 91K; refer to Table~\ref{tab:expResults2} for details.
For these experiments, we evaluate all three Resnet-18, 50 and 101 backbones and apply color and geometric augmentation during trainig.
We observed that a larger training set improves the performance of all the model significantly. 
For instance, training with full-dataset (91K images) improves the classification accuracy by an absolute 10.33\% (from 77.19\% to 87.52\%) compared to a model trained only with 10\% of the dataset (9K images) even for the small ResNet-18 model, thus benefiting from the large diversity in the dataset.

\begin{table}[ht]
\begin{center}
\begin{tabular}{l c c c c c}
\toprule
Dataset  & Images  & \multicolumn{3}{c}{Model Accuracy} \\
Size & & R-18 & R-50 & R-100\\
\toprule
Ours (10\%) & 9K & 77.19 & 79.65 & 79.84   \\
Ours (20\%) & 18K & 80.2 & 82.96 & 82.93  \\
Ours (50\%) & 45K & 85.59 & 85.55 & 76.36 \\
Ours (100\%) & 91K & \textbf{87.52} & \textbf{88.43} & \textbf{87.97}  \\
\toprule
\end{tabular}
\end{center}
\caption{Ablation with variations in number of training images w.r.t. our full corpus of 91K images. Using all available data achieves the best performance. Here, R=ResNet.} %
\label{tab:expResults2}
\end{table}

%% file: pages/06_conclusion.tex
\section{Conclusion and Future Work}

In this work, we have introduced the first large-scale dataset for Indian ethnic cloth classification. Our dataset is diverse consisting of images gathered from a large number of Indian e-commerce websites. 
We train several image classification models to benchmark our dataset.
We hope that our new dataset, which we will publish along with this work, will lead to the development of several algorithms (cloth classification, clothing landmark detection, etc) specifically for ethnic clothes.

Even though we achieve 88.43\% of classification accuracy, more research towards developing better classification models can incentivize cloth categorization. The dataset can be improved by balancing the number of images per category. In future, we plan to extend the dataset by introducing more categories from other ethnic groups.

\clearpage